\documentclass{article}

\PassOptionsToPackage{numbers, sort&compress}{natbib}
\usepackage[final]{neurips_2025}

\usepackage{microtype}
\usepackage{graphicx}
\usepackage{subfigure}
\usepackage{booktabs} 
\usepackage{xcolor}
\usepackage{colortbl}
\usepackage{multirow}

\definecolor{DarkGreen}{RGB}{1,200,32}
\definecolor{DarkRed}{RGB}{200,0,0}
\usepackage{hyperref}
\usepackage{wrapfig}

\hypersetup{
    colorlinks = true,
    citecolor=cyan,
    linkcolor=cyan,
    urlcolor=cyan
}

\usepackage{amsmath}
\usepackage{amssymb}
\usepackage{mathtools}
\usepackage{amsthm}
\usepackage{bm}

\usepackage[capitalize,noabbrev]{cleveref}

\theoremstyle{plain}

\theoremstyle{definition}

\theoremstyle{remark}

\usepackage[utf8]{inputenc} 
\usepackage[T1]{fontenc}    
\usepackage{hyperref}       
\usepackage{url}            
\usepackage{booktabs}       
\usepackage{amsfonts}       
\usepackage{nicefrac}       
\usepackage{microtype}      
\usepackage{graphicx}
\usepackage{subfigure}
\usepackage{booktabs}
\usepackage{colortbl}
\usepackage{multirow}

\usepackage[small, bf]{caption}
\usepackage[inline]{enumitem}

\title{Activation-Informed Merging of\\ Large Language Models}

\author{%
  Amin Heyrani Nobari\thanks{Equal Contribution.} ~\textsuperscript{1}, 
   Kaveh Alim$^*$\textsuperscript{1},
  Ali ArjomandBigdeli\textsuperscript{2} \\ 
  \textbf{Akash Srivastava\textsuperscript{3},
  Faez Ahmed\textsuperscript{1},
  Navid Azizan\textsuperscript{1}} \\[2mm]
  \textsuperscript{1}Massachusetts Institute of Technology \\
  \textsuperscript{2}Stony Brook University \\ 
    \textsuperscript{3}MIT-IBM Watson AI Lab \& Red Hat AI Innovation\\
}

\begin{document}

\maketitle

\begin{abstract}

Model merging, a method that combines the parameters and embeddings of multiple fine-tuned large language models (LLMs), offers a promising approach to enhance model performance across various tasks while maintaining computational efficiency. 
This paper introduces Activation-Informed Merging (AIM), a technique that integrates the information from the activation space of LLMs into the merging process to improve performance and robustness. 
AIM is designed as a flexible, complementary solution that is applicable to any existing merging method.
It aims to preserve critical weights from the base model, drawing on principles from continual learning~(CL) and model compression.
Utilizing a task-agnostic calibration set, AIM selectively prioritizes essential weights during merging. We empirically demonstrate that AIM significantly enhances the performance of merged models across multiple benchmarks. Our findings suggest that considering the activation-space information can provide substantial advancements in the model merging strategies for LLMs with up to 40\% increase in benchmark performance. Our code is publicly available at \href{https://github.com/ahnobari/ActivationInformedMerging}{https://github.com/ahnobari/ActivationInformedMerging}.
\end{abstract}

\section{Introduction}
\label{sec:intro}
Foundation models are rapidly becoming the dominant force for building Artificial Intelligence~(AI) systems. In many cases, researchers build their machine learning models by starting from pre-trained foundation models and fine-tuning~(FT) these pre-trained models for some desired target task \citep{pareja2024}. In such a paradigm, numerous fine-tuned models are developed for various tasks. However, an important opportunity is missed, as these fine-tuned task-specialized models typically operate in isolation without leveraging the rich features that each possesses \citep{sudalairaj2024lab}. This fact highlights the importance of a growing area of research focused on combining multiple task-specialized models fine-tuned from the same base foundation model.

In particular, as large language models (LLMs) continue to evolve, it becomes increasingly important to develop methods that can effectively fuse the specialized knowledge of various fine-tuned models derived from the same foundation model. Model merging has shown broad applications, including enhancing accuracy and robustness \citep{ModelSoups}, improving generalization \citep{rame2023ratatouille}, multi-modal models \citep{sung2023multimodalmerge}, and model alignment to human feedback \citep{WARM, WARP}. Given these benefits, a substantial amount of attention has been devoted to developing more effective merging algorithms for LLMs. 

In the vast majority of cases, merging LLMs is done using algorithms that explore the weight space of models and do not leverage the information in the activation space. Activation space information has been widely used to develop model pruning and compression methods, both in the context of general deep learning methods \citep{frantar2022OBS}, and more specifically for large language models \cite{lin2024awqactivationawareweightquantization}. However, this direction has remained underexplored for developing more robust merging algorithms. We hypothesize that the activation space information may hold key insights useful for model merging and explore this in our work.

\begin{figure*}[t]
    \centering
    \includegraphics[width=0.7\linewidth]{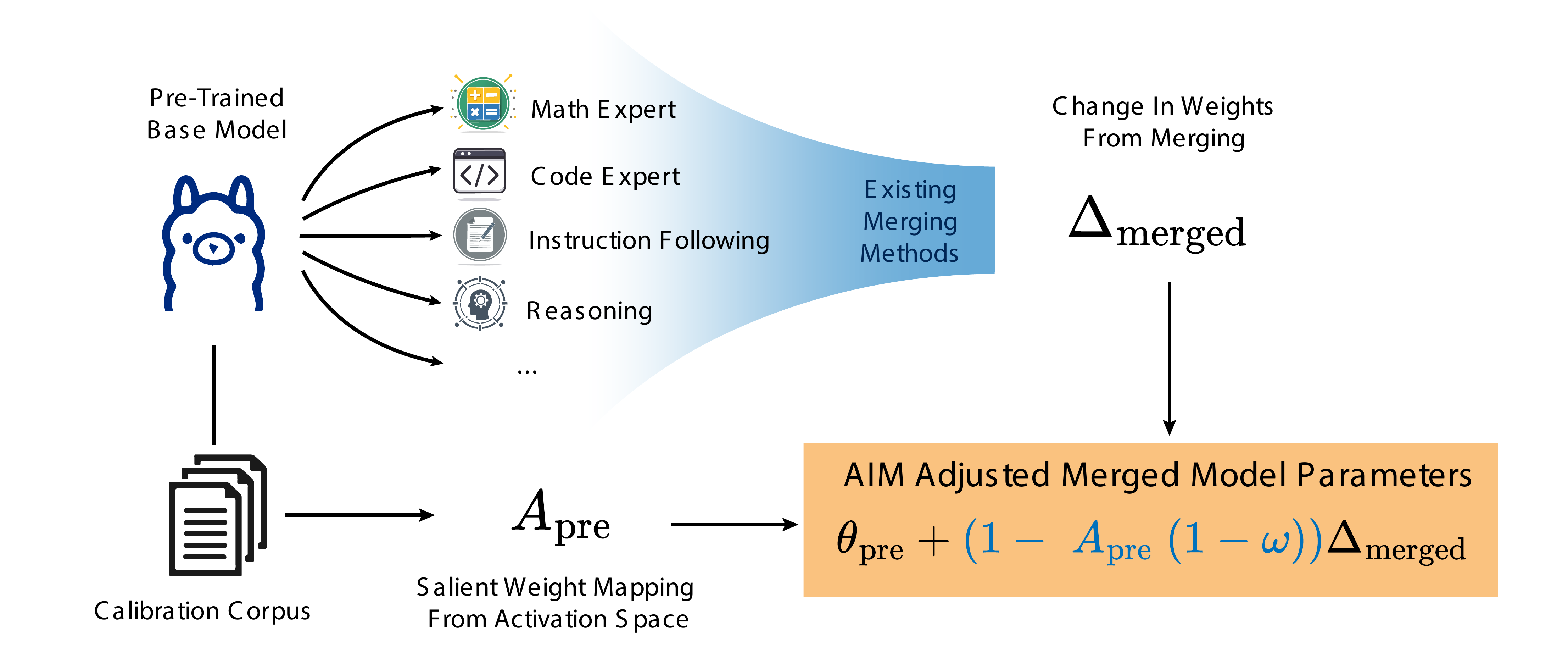}
\caption{Overview of the proposed activation-informed merging (AIM) in LLMs.}
    \label{fig:overview}
    \vspace{-2em}
\end{figure*}

In this paper, we view the problem of merging from a continual learning perspective. Specifically, we explore how FT models and merging them can significantly deviate from the pre-trained base model, potentially leading to overall performance degradation. This is analogous to the common catastrophic forgetting problem in continual learning~\cite{farajtabar2020OGD, Wang2024CLSurvey, sudalairaj2024lab}. Given this perspective, we incorporate the activation space information, taking into account the importance of the base model in preserving its pre-trained capabilities while integrating new knowledge from fine-tuned models. 
To achieve this, we introduce a new method named Activation-Informed Merging (AIM), which modifies the update step in the merging process to ensure that the most influential weights of the base model, identified through its activations, undergo minimal changes.
Figure \ref{fig:overview} demonstrates this basic process of AIM.

AIM fundamentally relates to the widely used approach of weight regularization in continual learning \citep{kirkpatrick2017overcomingCatastrophic, ritter2018laplaceCatastrophic, aljundi2018memory}.  When merging large language models fine-tuned from the same base model, the goal is to maintain the base model's performance while improving the merged model's expertise using the fine-tuned models. 
Various methods have been proposed to merge fine-tuned LLMs \citep{widen, DARE, TaskArithmetic, ModelSoups, davari2024modelbreadcrumbsscalingmultitask, yadav2023tiesmergingresolvinginterferencemerging, FisherMerging, jin2023dataless}. Although useful, these methods are fragile to outliers and low-quality fine-tuned models and may perform worse than the base model. Hence, we take a new perspective toward merging and adopt the continual learning view to prevent catastrophic forgetting.
Our extensive experimental study shows that AIM is a complementary solution that can be applied in conjunction with \emph{any} prior merging methods and consistently improves their performances across all tested benchmarks—including math, code, and instruction following—by up to 40\%. Despite its simplicity, AIM shows the effectiveness and importance of the activation space information for more effective model merging. 
\vspace{-1em}

\section{Background and Related Work}
\label{sec:background}

\paragraph{LLM Merging.}
Fine-tuning pre-trained language models has become the prevalent paradigm for building downstream NLP models. Often, fine-tuned models are readily available, but their training data is not due to data privacy or intellectual property concerns \citep{dodge2020finetuningpretrainedlanguagemodels}. With many fine-tuned checkpoints from the same pre-trained model, LLM merging has emerged as a complementary approach to fine-tuning, combining multiple task-specialized models into a unified architecture. This technique offers several advantages: it reduces storage and inference costs by consolidating expertise into a single model, facilitates compositional generalization by integrating specialized capabilities from different fine-tuned models, and supports decentralized model development by allowing independently trained models to be merged efficiently \citep{yadav2024mattersmodelmergingscale}.

Model soup, proposed by \citet{ModelSoups}, demonstrates that even simple averaging of the weights of multiple fine-tuned models can enhance accuracy and robustness in image classification and natural language processing applications. Further, \citet{rame2023ratatouille} show that model merging can improve out-of-distribution generalization performance. Extending this approach beyond unimodal settings, \citet{sung2023multimodalmerge} empirically demonstrate that model merging is also effective in multimodal setups. Beyond generalization benefits, model merging has also been explored for alignment; WARP \citep{WARP} and WARM \citep{WARM} introduce weight merging strategies to improve alignment in reinforcement learning from human feedback. WARP demonstrates that merging rewarded policies enhances model quality and alignment, while WARM shows that weight-averaged reward models improve robustness and help mitigate reward hacking.

Recently, many methods have been proposed to go beyond simple weight averaging to merge fine-tuned LLMs into a multitask model by combining their capabilities. Spherical Linear intERPolation (SLERP) \citep{Slerp}, originally proposed for animating rotations, which interpolates between two checkpoints, can be seen as a modification to simple weight averaging that finds a spherical path instead of a linear path in models' parameter space. A main shortcoming of this approach is that it does not support merging more than two models. \citet{ModelStock} also leverages geometric insights, showing that merging only two fine-tuned models can provide superior in-distribution and out-of-distribution performance compared to ensembling multiple models.

Task Arithmetic \citep{TaskArithmetic} generalizes simple weight averaging by introducing task vectors. It suggests moving the base model parameters in the direction of a weighted average of fine-tuned model differences with respect to the base model. Followed by the introduction of task vectors, Model Breadcrumbs \citep{davari2024modelbreadcrumbsscalingmultitask}, Trim, Elect Sign \& Merge (TIES merging)  \citep{yadav2023tiesmergingresolvinginterferencemerging}, and Drop And REscale (DARE) \citep{DARE} leverage pruning techniques for better and more scalable ways of merging task vectors. WeIght DisENtanglement based merging (WIDEN) \citep{widen} takes a more sophisticated approach to model merging by disentangling and analyzing weight components. 
Another line of work on model merging takes advantage of the information in the model activations of the training data. \citet{FisherMerging} propose Fisher merging, which leverages the Laplace approximation by using the diagonal of each model’s Fisher information. \citet{jin2023dataless} attempt to minimize prediction differences between the merged model and the individual models and introduce the Regression Mean (RegMean) method that calculates the optimal weights with respect to Euclidean distance to model predictions.

We refer the reader to \citet{yang2024modelmergingllmsmllms} for a more comprehensive literature overview. The study provides a detailed discussion of model merging methods and theories, explores their applications in LLMs and multimodal large language models, and highlights future research directions.

\paragraph{Continual Learning.}
Continual learning strategies can be categorized into five overarching approaches: regularization, where parameters are constrained or updated based on past data; replay, where past data is replayed for the model as it encounters new data; optimization, where the loss function and optimizer are targeted; representation, where new data representations and learned embeddings can be exploited for less forgetting; and architecture, where models and parameters can expand as new data arrives \cite{Wang2024CLSurvey}. 
To avoid catastrophic forgetting of the base model's abilities, we view the model merging problem through the lens of CL, primarily focusing on regularization-based methods.  Regularization-based methods penalize deviation from the base model according to some norm \citep{shi2024continualLLM}. Various methods have been proposed to mitigate catastrophic forgetting, e.g., \citet{aljundi2018memory, kirkpatrick2017overcomingCatastrophic, ritter2018laplaceCatastrophic}. 
In particular, Elastic Weight Consolidation (EWC) employs a Fisher Information Matrix to identify and protect parameters crucial for previous tasks by adding a quadratic penalty on the parameter shifts. 
Integrating CL approaches into the merging framework involves defining a weighted regularization term that selectively constrains parameter updates in critical areas for retaining previously learned tasks. 
This integration not only mitigates the risk of catastrophic forgetting but also enhances the adaptability and utility of the merged model.

\paragraph{Model Compression.}
Using activation space information has been shown to be useful in the context of model compression. \citet{frantar2022OBS} show that using a calibration dataset, deep learning models can be quantized and/or pruned efficiently. \cite{lin2024awqactivationawareweightquantization} introduce Activation-aware Weight Quantization (AWQ) for LLM compression and show that protecting only 1\% salient weights can greatly reduce quantization error. 

Building on these ideas, this paper presents a complementary merging approach that utilizes base model activations and principles from AWQ. Our method efficiently sketches delta parameters, ensuring the base model retains its original capabilities while incorporating expertise from fine-tuned models.

\section{Methodology: Activation-Informed Merging}
\label{sec:methodology}
As discussed in Section \ref{sec:background}, most existing approaches for merging FT LLMs primarily focus on the weight space of the models being merged. However, it is well established that the activation space of these models contains crucial insights into the degree of importance of different parameters of LLMs. This was shown to be the case, for instance, in the work done by \citet{lin2024awqactivationawareweightquantization} on Activation-aware Weight Quantization~(AWQ), outperforming traditional quantization methods by including insight from the activation space of LLMs. Given this, we hypothesize that the activation space of LLMs likely holds useful clues for model merging as well. Inspired by AWQ, we introduce Activation-Informed Merging~(AIM) for merging FT LLMs. In this section, we detail our proposed solution and discuss some of the inner workings of AIM.

\subsection{The Merging Problem and Connections to Continual Learning}

Consider the merging of $N$ models with parameters ${\theta}_1, {\theta}_2, \cdots, {\theta}_N$ fine-tuned on different tasks from a common pre-trained model with parameters ${\theta}_{\text{pre}}$. For each fine-tuned LLM, we are essentially creating experts on specific tasks that move away from the generalist pre-trained model, hence usually degrading performance in some tasks while improving performance on the task for which the model is fine-tuned. In this sense, each FT model with parameters ${\theta}_n$ can be seen as a model fitted to a new task \( \mathcal{D}_n = \{X_n, Y_n\} \) in a continual learning scenario with the potential for catastrophic forgetting on the generalist pre-trained model, which may not perform as well on the specific task but will have a more balanced performance across various tasks. As such, we hypothesize that when merging FT LLMs adapted from the same base model, emphasis on the base model can build better robustness to large performance degradation across numerous tasks while still allowing for capturing each FT expert's capabilities. AIM seeks to achieve this by relaxing the changes to the salient weights of the base model in the final merged model. In this way, AIM is analogous to weight regularization in many continual learning approaches~\cite{Wang2024CLSurvey, ritter2018laplaceCatastrophic, kirkpatrick2017overcomingCatastrophic, aljundi2018memory, shi2024continualLLM}.
Notably, the saliency of weights is determined by analyzing the activation space (sensitivities) of the base model rather than just regularizing based on the weight space, similar to \cite{farajtabar2020OGD,min2022one}.
 
\subsection{Activation Space Analysis}

AIM determines the saliency of a model's weights by looking at the scale of activations by passing a calibration dataset to the model and recording the scale of activations in each channel. To better understand why, we will analyze how perturbation of the weights of a given model affects the model outputs. Let the original weights be \( w \in \mathbb{R}^{N\times M}\) and the perturbation be \( \delta w \in \mathbb{R}^{N\times M}\), such that the perturbed weights are \(w' = w + \delta w\). The output of a linear layer with input \( x \in \mathbb{R}^{1 \times N}\) and perturbed weights is
\begin{equation}
y' = xw' = x(w + \delta w) = xw + x(\delta w).
\end{equation}
The magnitude of the error due to this perturbation, \(\text{Error} = y' - y = x(\delta w)\), scales with the magnitude of activation \( x \):
\begin{equation}
\|\text{Error}\|_p = \|\operatorname{diag}(x)  \delta w\|_p \leq \|\operatorname{diag}(x) \delta w\|_1  \leq
\sum_{i=1}^{N}|x_i|\sum_{j=1}^{M}|\delta w_{ij}|.
\end{equation}

For any specific input channel \( x_i \), the error contribution from perturbation in the $i$-th row of $w$, \( \delta w_i \) is amplified by the magnitude of the same channel in the input. As such, one could selectively regularize the weights based on the importance of the input channels, i.e., their magnitudes. In this way, we use a calibration dataset to capture the average magnitude of the input channels for each layer of the base model and determine the saliency of weights in the base model from the activation space.

\subsection{Calibration Data and Robustness to Calibration Data}
We choose the calibration dataset to be a subset of the validation data from the pile dataset~\cite{gao2020pile800gbdatasetdiverse}, which is similar in distribution to most pre-training datasets. Most notably, this dataset has been utilized in model compression for both quantization in AWQ~\cite{lin2024awqactivationawareweightquantization} and pruning in WANDA~\cite{sun2024simpleeffectivepruningapproach}. This calibration dataset is considered to be fairly diverse and not task-specific, which should allow us to quantify weights' saliency in a robust manner.

\paragraph{Robustness to Calibration Data} The use of activation spaces from calibration data has been thoroughly studied in model compression, by \citet{lin2024awqactivationawareweightquantization} and \citet{sun2024simpleeffectivepruningapproach}. In both cases, authors conduct extensive studies on the robustness of their methods, which, like ours, rely on the magnitude of activations, and both concur in two important findings. 1) When using calibration data to quantify activation space magnitudes to be used by algorithms for model quantization or pruning, performance is robust to dataset quality. This is in contrast to methods that require fine-tuning or retraining after compression. This observation is made clear in both studies and confirms that sensitivity to the quality of the data is much less in methods that require only activation space analysis without training. 2) Most notably, ablation studies on the size of the calibration set also show robustness to the size of calibration data, with both studies confirming that only 8-32 sample sequences of length approximately 2048 tokens are enough for model compression algorithms to produce robust outcomes~\cite{lin2024awqactivationawareweightquantization,sun2024simpleeffectivepruningapproach}. Despite this robustness, we use the same 256 total sequences (approximately 524K tokens) that the authors of both studies use in their main experiments; however, this robustness to dataset size is noted as an important tool for reducing computational cost and time for running algorithms such as ours.
We also study this matter in an ablation study and show that, like those studies, AIM is also robust to dataset size and can be made much more efficient if need be.

\subsection{Adaptable Relaxation Scheme}
As discussed, we introduce an adaptable relaxation scheme based on the activations of the base model, which we wish not to stray away from significantly. To make the scheme adaptable to any merging algorithm in the weight space, we formulate our relaxation scheme in terms of the changes in the weights. Given a task-agnostic representative calibration corpus $D$. We can pass this corpus through the model and accumulate the activations from each token in the dataset. This will yield the average magnitude of activations for each channel in all layers of the model. As previously noted, averaging the inputs over the calibration set yields a vector $x \in \mathbb{R}^{1\times N}$ for each linear layer. Constructing a diagonal matrix from the absolute values of this vector gives $ \operatorname{diag}(|x|) \in \mathbb{R}^{N \times N}$. By normalizing $\operatorname{diag}(|x|)$ using its maximum value, we obtain a diagonal matrix $A_{\text{pre}} \in [0, 1]^{N \times N}$. This matrix serves to modulate weight updates based on input saliency, allowing for consistent control of merging through a universal hyperparameter $\omega$. 
Next, we define the action of any given merging method by the changes that it applies to the model weights with respect to each of the fine-tuned models being merged~(i.e., ${\theta}_1, {\theta}_2, \ldots, {\theta}_n$). Specifically, we denote the weight update contributed by a fine-tuned model (with parameters ${\theta}_i$) to the model parameters by ${\Delta}_i$~(e.g., ${\Delta}_i = \frac{{\theta}_{i}-{\theta}_\text{pre}}{n}$ for weight averaging). Now, we propose an adaptive relaxation scheme that adjusts the final model. For simplicity of notation, let ${\theta}$ refer to weights of a linear layer (i.e., an $N\times M$ matrix), then the AIM relaxation scheme can be written as:

\vspace{-1.5em}
\begin{equation}
    \delta w_{\text{AIM}}={\theta}_\text{merged} - {\theta}_\text{pre} = (1-A_\text{pre}(1-\omega))\sum_{i=1}^{N} \lambda_i {\Delta}_i,
    \label{eqn:AIM}
\end{equation}
\vspace{-1em}

where $\delta w_{\text{AIM}}$ is the relaxation change and the subscript \textit{pre} refers to the pre-trained model and $\omega$ is the relaxation factor that controls how much relaxation is applied (an $\omega$ of 0.0 would revert the most salient weight to the base weights and an $\omega$ of 1.0 applies no relaxation), effectively $\omega$ is scaling error/deviation from the base model, and $\lambda_i$ are the weight factors for each $\Delta_i$. Note that $\lambda_i$ is internal to each merging algorithm and not part of the AIM relaxation; however, in general, one could selectively apply this if desired. In this work, since we do not explicitly look at the merging algorithm's inner workings, we can simply fuse the terms \(\sum_{i=1}^{N} \lambda_i {\Delta}_i\) into a single algorithm-agnostic term \(\Delta_{\text{merged}}\) and simplify the relaxation scheme to:
\begin{equation}
    {\theta}_\text{AIM} = {\theta}_\text{pre} + (1-A_\text{pre}(1-\omega))\Delta_{\text{merged}}.
    \label{eqn:AIM_simp}
\end{equation}

Note that in general $A_\text{pre}$ is not a single matrix, rather a mapping of activations for all model parameters obtained as we described prior. In our experiments, we apply this relaxation scheme to several different merging methods and explore how the hyperparameter $\omega$ affects the merged model's behavior, and present these results in Section \ref{sec:results}.

\paragraph{Sensitivity-Based Formulation}
An alternative way of choosing the importance scores for changing the model weights---which has been used, e.g., in continual learning~\cite{farajtabar2020OGD,min2022one}, out-of-distribution detection~\cite{sharma2021sketching}, and meta-learning~\cite{almecija2022uncertainty}---is to use the partial derivatives of the (base) model with respect to the parameters, i.e., \emph{sensitivities}, which correspond to a scaled version of the activations. This is in contrast with using the activations directly, common in model compression. Formally, let $f_{{\theta}}$ denote the model with parameters ${\theta}$, and $\theta[j]$ be one of the weights; then $g_{\theta[j]}=|\frac{\partial f}{\partial \theta[j]}|$ determines how sensitive the output is to perturbing the weight $\theta[j]$. We develop a sensitivity-based formulation of AIM by incorporating sensitivity scores calculated from the gradients and replacing activations with sensitivity scores. We consider $f_{{\theta}}(x)$ to be the logits of the model with parameter $\theta$ and $\mathcal{L}$ to be the entropy function, similar to \citet{farajtabar2020OGD}'s suggested approach for classification problems. 
Now let $G_\text{pre} = |\frac{\partial \mathcal{L}}{\partial {\theta}}|$ be the magnitude of the gradient for all parameters of the pretrained model for samples in a calibration corpus $D$. Then the sensitivity-based formulation can be written as follows:

\begin{equation}
    {\theta}_{\text{AIM}, \text{G}} = {\theta}_\text{pre} + (1-G_\text{pre}(1-\omega))\Delta_{\text{merged}}.
    \label{eqn:AIMG}
\end{equation}

In other words, the sensitivity-based formulation of the problem suggests that regularization of weights through relaxation should be done based on model gradients. However, we note that computing gradients for a calibration corpus can be computationally expensive and will require significantly more memory and compute resources than storing activations, which is akin to performing inference on the model. To verify that using activations retains the same performance and fidelity as this formulation, we conduct our experiments for both AIM and the sensitivity-based formulation. We present the results of the continual learning view in Appendix~\ref{app:continualexp}, and we see that when comparing results of AIM (Table \ref{tab:results}) and the sensitivity-based formulation (Table \ref{tab:continual}), the performance boost of both approaches is very similar, with AIM being computationally more efficient. 

\section{Experimental Setup and Evaluation Metrics}
\label{sec:experiments_metrics}
We conduct two separate experiments with AIM: 1) we apply AIM to 5 different merging methods including the two latest works on the topic with different numbers of experts being merged and report the performance of the models on 6 different benchmarks; 2) we conduct an ablation study on the $\omega$ parameter in AIM and analyze how $\omega$ affects each of the merging methods in a scenario where 3 different experts are being merged.

\subsection{Selection of FT Expert LLMs}
To understand how AIM reacts with different merging methods, we conduct experiments with merging different experts fine-tuned from the same base model. The set of experts we use is the same set of experts used by the two latest LLM merging algorithms in the literature, namely DARE~\cite{DARE} and WIDEN~\cite{widen}, which use the same three experts fine-tuned from Llama-2-13b~\cite{llama2}. These experts include the WizardLM-13B~\cite{xu2024wizardlm} model fine-tuned for instruction following, WizardMath-13B~\cite{wizardmath} fine-tuned for superior mathematical reasoning, and llama-2-13b-code-alpaca, which serves as the code expert~\cite{codealpaca}. See Appendix~\ref{app:reporducibility} for more details.

\subsection{Merging Methods Implementations}
In our experiments, we implement the latest merging methods in the literature for LLM merging. These include newly developed DARE and WIDEN~\cite{DARE,widen} methods as well as some of the long-established approaches of TIES merging~\cite{yadav2023tiesmergingresolvinginterferencemerging}, and task arithmetic~\cite{TaskArithmetic}. For all merging methods except WIDEN, we use the comprehensive MergeKit implementations developed by \citet{goddard-etal-2024-arcees}, and for WIDEN, we use the publicly available implementation provided by the authors of the paper.

We note that in many of the merging algorithms, many hyperparameters can be adjusted. In these cases, we use the author-recommended values where available and the default parameters recommended by \citet{goddard-etal-2024-arcees}. Note that it is possible to perform a grid search on these hyperparameters to find optimal values for each benchmark; however, this would essentially be over-fitting on benchmarks and does not provide any value to our analysis of the proposed complementary relaxation scheme, which applies adjustments to the merged models. For reproducibility, all of our checkpoints and code to reproduce the results will be made publicly available.

\subsection{Benchmarks Used For Evaluations}
Given that the expert models we use in our experiments involve fine-tuning on instruction following, mathematical reasoning, and code generation, we use several common benchmarks for each of these tasks. Specifically, we measure model performance on language understanding with the MMLU~\cite{hendrycks2020measuring} benchmark, instruction following with IFEval~\cite{zhou2023instructionfollowingevaluationlargelanguage} benchmark, code generation with HumanEval~\cite{chen2021codex} and MBPP~\cite{austin2021program} benchmarks, and mathematical reasoning with the MATH~\cite{hendrycks2021measuringmathematicalproblemsolving} and GSM8K~\cite{gsm8k} benchmarks. For all benchmarks, we use the latest versions and up-to-date implementations developed by \citet{lmharness} except for mathematical reasoning, for which we use the chain of thought prompting used by \citet{wizardmath} to replicate the results of the original model as closely as possible. The code we use for these benchmark results will also be made publicly available for reproducibility.

In addition to the common benchmarks that we use to evaluate merged models, we also propose a new evaluation metric for LLM merging~(or merging of different experts in general), which we believe helps better contextualize the value added by any given merging algorithm and which we discuss in the following section.

\subsection{Measuring Performance From an Optimization Perspective}
\label{subsec:hvmetric}
We note that in most cases, LLMs are not meant to operate as narrow expert models, unlike a large portion of deep learning applications where models are trained to perform very specific tasks such as classification or regression. LLMs, in contrast, are generalist language models aiming to assist across a wide variety of tasks and applications. As such, LLMs can be viewed from a multi-objective optimization perspective. In merging scenarios specifically, multiple expert models are brought together to create a merged model aiming to find the balance of performance across the different expertise of the fine-tuned models. In this sense, each expert can be thought of as optimized for a specific objective. This perspective lends itself rather well to a multi-objective optimization view of the problem. Given this, only looking at how each model performs on each of the benchmarks does not give us a full picture of the multi-objective goal of merging.

To obtain a more comprehensive view of merging performance, we propose a \textbf{hypervolume-based metric} that quantifies the contribution of the merged model to the \textbf{multi-objective frontier} of FT LLMs. Consider a performance space defined over $N$ benchmarks, where each model's performance is represented as a point in an $N$-dimensional space. The performance on each benchmark is normalized to the range $[0,1]$, where $0$ represents the worst-case (reference point), and $1$ represents the best possible performance with 100\% accuracy on the benchmark in question.

Let $\mathbf{r} = (r_1, r_2, \dots, r_N)$ denote the reference point in this space, which we set to $(0,0,\dots,0)$ to ensure hypervolume calculations are consistently defined. Given a set $S$ of FT models and the pre-trained base model, let $S^* \subseteq S$ denote the subset of models that are \textbf{Pareto-optimal}, i.e., models that are not dominated by any other model in $S$. The hypervolume of this set, denoted as $\operatorname{HV}(S^*)$, is defined as:

\vspace{-1em}
\begin{equation}
    \operatorname{HV}(S^*) =  \lambda \left( \bigcup_{\mathbf{x} \in S^*} \text{dom}(\mathbf{x}, \mathbf{r}) \right),
\end{equation}
\vspace{-1em}

where $\lambda(\cdot)$ denotes the Lebesgue measure (i.e., volume in $\mathbb{R}^N$), and $\text{dom}(\mathbf{x}, \mathbf{r})$ represents the hypervolume dominated by $\mathbf{x}$ with respect to the reference point $\mathbf{r}$.

When a merged model $M$ is introduced, the new set becomes $S' = S \cup \{M\}$, and the updated Pareto-optimal set is denoted as $S'^*$. Given this set including the merged model, we can measure the added value of the merged model from a multi-objective perspective as the normalized hypervolume gain~(HV Gain) as a result of adding this merged model:

\vspace{-1em}
\begin{equation}
    \text{HV Gain} = \sqrt[d]{\operatorname{HV}(S'^*) - \operatorname{HV}(S^*)},
\end{equation}

where $d$ is the number of dimensions/benchmarks. Since hypervolume is computed only over Pareto-optimal models, we have $\operatorname{HV}(S'^*) \geq \operatorname{HV}(S^*)$, ensuring that $\text{HV Gain} \geq 0$. This metric provides an aggregated measure of merging effectiveness, capturing trade-offs across multiple benchmarks rather than focusing on isolated improvements, thus providing a full picture of merging performance. In our experiments, we track \textit{HV Gain} along with the 6 aforementioned benchmarks.

\section{Experiments and Results}
 
\label{sec:results}
In this section, we present our results on applying AIM to different merging methods as well as an ablation study on how the $\omega$ hyperparameter affects the performance of each merging method. All experiments are run using 4 H100 GPUs, and each set of benchmarks takes roughly 15 minutes to run for each checkpoint.

\subsection{AIM Applied to Various Merging Approaches}

\begin{figure*}
    \centering
    \includegraphics[width=0.8\linewidth]{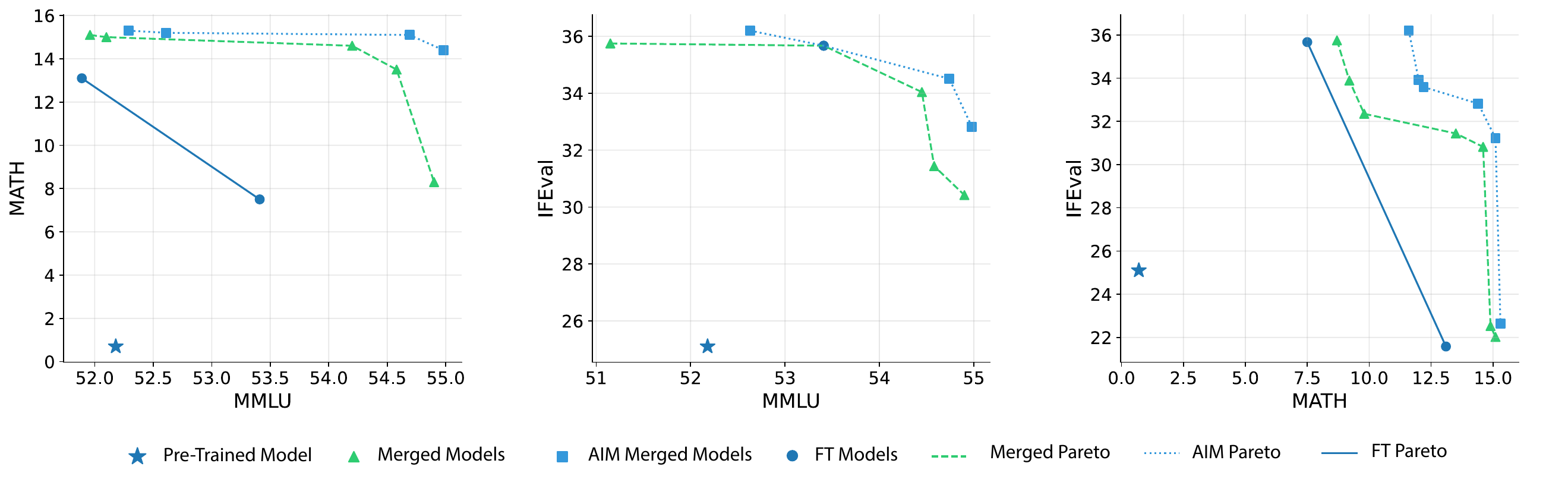}

    \caption{The Pareto fronts of models under different scenarios. Note that the points in these plots represent all models benchmarked in Table \ref{tab:results}, for better readability, we only visualize the dominating points in each case. The measured increases in HV Gain when AIM is applied can be clearly seen in the Pareto frontier shifting further forward when AIM is applied compared to when only a population of merged models is evaluated.}
    \label{fig:frontier}
    \vspace{-1em}
\end{figure*}

\begin{table*}[ht]
\centering
\caption{
Benchmark Results Across Various Merging Methods. Percentage changes are shown relative to models merged without AIM. The highest-performing fine-tuned and base models are highlighted in yellow, and the best-performing merged models are marked in blue. The results demonstrate that applying AIM significantly enhances the performance of merged models.}

\small
\resizebox{\textwidth}{!}{
\begin{tabular}{l|l|l|cccccc|c}
\hline
Method & Model(s) & AIM & HumanEval & MBPP & MMLU & MATH & GSM8K & IFEval & HV Gain \\
\hline
\multicolumn{9}{c}{\textbf{Base Models}} \\
\hline
- & Base & - & 17.07 & 27.80 & 52.18 & 0.70 & 4.20 & 25.10 & - \\
- & Code & - & 17.07 & 31.60 & 52.91 & 6.00 & 24.10 & 26.25 & - \\
- & Instruction Tuned & - & \cellcolor{orange!20}{26.83} & \cellcolor{orange!20}{34.80} & \cellcolor{orange!20}{53.41} & 7.50 & 43.40 & \cellcolor{orange!20}{35.67} & - \\
- & Math & - & 15.24 & 27.60 & 51.89 & \cellcolor{orange!20}{13.10} & \cellcolor{orange!20}{59.10} & 21.58 & - \\
\hline
\multicolumn{9}{c}{\textbf{Merged Models}} \\
\hline
\multirow{8}{*}{DARE Task Arithmetic} & \multirow{2}{*}{Code + Instruction Tuned} & No & 26.83 & 34.40 & 53.53 & 8.40 & 45.80 & 33.42 & 0.27 \\
& & Yes & 29.27 \color{DarkGreen}(+9.09\%) & 36.00 \color{DarkGreen}(+4.65\%) & 54.18 \color{DarkGreen}(+1.21\%) & 8.30 \color{DarkRed}(-1.19\%) & 46.20 \color{DarkGreen}(+0.87\%) & 32.00 \color{DarkRed}(-4.25\%) & 0.28 \color{DarkGreen}(+2.49\%) \\
 & \multirow{2}{*}{Code + Math} & No & 16.46 & 28.60 & 51.96 & 15.10 & 64.70 & 22.02 & 0.23 \\
& & Yes & 15.85 \color{DarkRed}(-3.71\%) & 29.60 \color{DarkGreen}(+3.50\%) & 52.50 \color{DarkGreen}(+1.04\%) & 14.80 \color{DarkRed}(-1.99\%) & 64.10 \color{DarkRed}(-0.93\%) & 21.91 \color{DarkRed}(-0.50\%) & 0.23 \color{DarkRed}(-1.65\%) \\
 & \multirow{2}{*}{Instruction Tuned + Math} & No & 5.49 & 19.00 & 51.08 & 9.80 & 54.30 & 32.35 & 0.18 \\
& & Yes & 12.20 \color{DarkGreen}(+122.22\%) & 28.20 \color{DarkGreen}(+48.42\%) & 52.72 \color{DarkGreen}(+3.21\%) & 12.90 \color{DarkGreen}(+31.63\%) & 62.20 \color{DarkGreen}(+14.55\%) & 31.96 \color{DarkRed}(-1.21\%) & 0.26 \color{DarkGreen}(+40.71\%) \\
 & \multirow{2}{*}{Code + Instruction Tuned + Math} & No & 11.59 & 19.60 & 50.89 & 9.10 & 49.70 & 33.20 & 0.16 \\
& & Yes & 15.85 \color{DarkGreen}(+36.76\%) & 27.00 \color{DarkGreen}(+37.76\%) & 52.59 \color{DarkGreen}(+3.34\%) & 12.20 \color{DarkGreen}(+34.07\%) & 60.70 \color{DarkGreen}(+22.13\%) & 33.59 \color{DarkGreen}(+1.17\%) & 0.23 \color{DarkGreen}(+40.59\%) \\
\hline
\multirow{8}{*}{DARE Ties} & \multirow{2}{*}{Code + Instruction Tuned} & No & 30.49 & 35.20 & 53.40 & 8.60 & 46.20 & 33.28 & 0.28 \\
& & Yes & \cellcolor{blue!20}{30.49} & \cellcolor{blue!20}{36.80} \color{DarkGreen}(+4.55\%) & 54.02 \color{DarkGreen}(+1.16\%) & 8.60 & 47.20 \color{DarkGreen}(+2.16\%) & 33.16 \color{DarkRed}(-0.36\%) & 0.29 \color{DarkGreen}(+1.63\%) \\
 & \multirow{2}{*}{Code + Math} & No & 17.07 & 27.40 & 51.92 & 14.90 & 63.60 & 22.53 & 0.23 \\
& & Yes & 17.68 \color{DarkGreen}(+3.57\%) & 29.00 \color{DarkGreen}(+5.84\%) & 52.61 \color{DarkGreen}(+1.33\%) & 15.20 \color{DarkGreen}(+2.01\%) & 63.90 \color{DarkGreen}(+0.47\%) & 21.10 \color{DarkRed}(-6.35\%) & 0.24 \color{DarkGreen}(+4.00\%) \\
 & \multirow{2}{*}{Instruction Tuned + Math} & No & 8.54 & 23.80 & 51.39 & 9.20 & 54.10 & 33.89 & 0.20 \\
& & Yes & 15.85 \color{DarkGreen}(+85.60\%) & 30.20 \color{DarkGreen}(+26.89\%) & 52.89 \color{DarkGreen}(+2.92\%) & 11.60 \color{DarkGreen}(+26.09\%) & 57.80 \color{DarkGreen}(+6.84\%) & 35.63 \color{DarkGreen}(+5.13\%) & 0.26 \color{DarkGreen}(+31.22\%) \\
 & \multirow{2}{*}{Code + Instruction Tuned + Math} & No & 13.41 & 21.20 & 51.15 & 8.70 & 51.50 & 35.75 & 0.17 \\
& & Yes & 19.51 \color{DarkGreen}(+45.49\%) & 28.60 \color{DarkGreen}(+34.91\%) & 52.63 \color{DarkGreen}(+2.89\%) & 11.60 \color{DarkGreen}(+33.33\%) & 57.00 \color{DarkGreen}(+10.68\%) & \cellcolor{blue!20}{36.20} \color{DarkGreen}(+1.26\%) & 0.24 \color{DarkGreen}(+41.28\%) \\
\hline
\multirow{8}{*}{Task Arithmetic} & \multirow{2}{*}{Code + Instruction Tuned} & No & 29.27 & 33.80 & 53.44 & 8.60 & 47.10 & 31.60 & 0.28 \\
& & Yes & 29.88 \color{DarkGreen}(+2.08\%) & 35.80 \color{DarkGreen}(+5.92\%) & 54.12 \color{DarkGreen}(+1.27\%) & 7.80 \color{DarkRed}(-9.30\%) & 46.60 \color{DarkRed}(-1.06\%) & 32.01 \color{DarkGreen}(+1.30\%) & 0.28 \color{DarkGreen}(+0.61\%) \\
 & \multirow{2}{*}{Code + Math} & No & 18.29 & 28.60 & 52.10 & 15.00 & 64.70 & 21.92 & 0.24 \\
& & Yes & 17.68 \color{DarkRed}(-3.34\%) & 29.20 \color{DarkGreen}(+2.10\%) & 52.52 \color{DarkGreen}(+0.81\%) & 14.60 \color{DarkRed}(-2.67\%) & 64.50 \color{DarkRed}(-0.31\%) & 21.54 \color{DarkRed}(-1.73\%) & 0.24 \color{DarkRed}(-2.65\%) \\
 & \multirow{2}{*}{Instruction Tuned + Math} & No & 4.27 & 20.20 & 51.50 & 10.00 & 54.20 & 31.31 & 0.18 \\
& & Yes & 8.54 \color{DarkGreen}(+100.00\%) & 26.40 \color{DarkGreen}(+30.69\%) & 52.83 \color{DarkGreen}(+2.58\%) & 12.80 \color{DarkGreen}(+28.00\%) & 61.30 \color{DarkGreen}(+13.10\%) & 32.62 \color{DarkGreen}(+4.18\%) & 0.24 \color{DarkGreen}(+34.52\%) \\
 & \multirow{2}{*}{Code + Instruction Tuned + Math} & No & 11.59 & 19.60 & 51.20 & 9.00 & 52.70 & 32.87 & 0.16 \\
& & Yes & 15.24 \color{DarkGreen}(+31.49\%) & 27.40 \color{DarkGreen}(+39.80\%) & 52.63 \color{DarkGreen}(+2.79\%) & 12.00 \color{DarkGreen}(+33.33\%) & 58.10 \color{DarkGreen}(+10.25\%) & 33.91 \color{DarkGreen}(+3.16\%) & 0.22 \color{DarkGreen}(+31.97\%) \\
\hline
\multirow{8}{*}{Ties Merging} & \multirow{2}{*}{Code + Instruction Tuned} & No & 16.46 & 23.60 & 52.70 & 2.70 & 5.40 & 24.48 & 0.00 \\
& & Yes & 15.24 \color{DarkRed}(-7.41\%) & 24.20 \color{DarkGreen}(+2.54\%) & 53.15 \color{DarkGreen}(+0.85\%) & 2.60 \color{DarkRed}(-3.70\%) & 5.20 \color{DarkRed}(-3.70\%) & 22.87 \color{DarkRed}(-6.58\%) & 0.05 \color{DarkGreen}(+inf\%) \\
& \multirow{2}{*}{Code + Math} & No & 15.85 & 26.80 & 51.86 & 14.30 & 62.60 & 21.63 & 0.20 \\
& & Yes & 15.85 & 28.60 \color{DarkGreen}(+6.72\%) & 52.29 \color{DarkGreen}(+0.83\%) & \cellcolor{blue!20}{15.30} \color{DarkGreen}(+6.99\%) & 63.80 \color{DarkGreen}(+1.92\%) & 22.64 \color{DarkGreen}(+4.67\%) & 0.23 \color{DarkGreen}(+13.55\%) \\
 & \multirow{2}{*}{Instruction Tuned + Math} & No & 28.05 & 34.60 & 54.45 & 8.70 & 44.70 & 34.04 & 0.23 \\
& & Yes & 27.44 \color{DarkRed}(-2.17\%) & 35.00 \color{DarkGreen}(+1.16\%) & 54.74 \color{DarkGreen}(+0.53\%) & 9.30 \color{DarkGreen}(+6.90\%) & 46.10 \color{DarkGreen}(+3.13\%) & 34.51 \color{DarkGreen}(+1.38\%) & 0.25 \color{DarkGreen}(+6.38\%) \\
 & \multirow{2}{*}{Code + Instruction Tuned + Math} & No & 21.34 & 29.20 & 53.97 & 6.30 & 29.20 & 26.95 & 0.11 \\
& & Yes & 20.73 \color{DarkRed}(-2.86\%) & 29.20 & 54.46 \color{DarkGreen}(+0.91\%) & 5.70 \color{DarkRed}(-9.52\%) & 23.70 \color{DarkRed}(-18.84\%) & 25.98 \color{DarkRed}(-3.60\%) & 0.11 \color{DarkGreen}(+4.33\%) \\
\hline
\multirow{8}{*}{WIDEN} & \multirow{2}{*}{Code + Instruction Tuned} & No & 26.22 & 35.60 & 54.90 & 8.30 & 45.00 & 30.42 & 0.27 \\
& & Yes & 25.61 \color{DarkRed}(-2.33\%) & 34.60 \color{DarkRed}(-2.81\%) & 54.97 \color{DarkGreen}(+0.13\%) & 8.20 \color{DarkRed}(-1.20\%) & 44.10 \color{DarkRed}(-2.00\%) & 31.60 \color{DarkGreen}(+3.88\%) & 0.26 \color{DarkRed}(-0.93\%) \\
 & \multirow{2}{*}{Code + Math} & No & 17.07 & 29.40 & 53.35 & 14.20 & 64.40 & 24.02 & 0.24 \\
& & Yes & 17.07 & 29.60 \color{DarkGreen}(+0.68\%) & 53.36 \color{DarkGreen}(+0.02\%) & 14.30 \color{DarkGreen}(+0.70\%) & 62.20 \color{DarkRed}(-3.42\%) & 23.95 \color{DarkRed}(-0.29\%) & 0.24 \color{DarkRed}(-1.22\%) \\
 & \multirow{2}{*}{Instruction Tuned + Math} & No & 24.39 & 30.40 & 54.20 & 14.60 & 66.00 & 30.82 & 0.30 \\
& & Yes & 23.78 \color{DarkRed}(-2.50\%) & 32.00 \color{DarkGreen}(+5.26\%) & 54.69 \color{DarkGreen}(+0.90\%) & 15.10 \color{DarkGreen}(+3.42\%) & \cellcolor{blue!20}{68.20} \color{DarkGreen}(+3.33\%) & 31.23 \color{DarkGreen}(+1.33\%) & \cellcolor{blue!20}{0.31} \color{DarkGreen}(+2.54\%) \\
 & \multirow{2}{*}{Code + Instruction Tuned + Math} & No & 25.00 & 33.20 & 54.58 & 13.50 & 64.20 & 31.44 & 0.29 \\
& & Yes & 26.83 \color{DarkGreen}(+7.32\%) & 32.80 \color{DarkRed}(-1.20\%) & \cellcolor{blue!20}{54.98} \color{DarkGreen}(+0.73\%) & 14.40 \color{DarkGreen}(+6.67\%) & 64.00 \color{DarkRed}(-0.31\%) & 32.82 \color{DarkGreen}(+4.39\%) & 0.30 \color{DarkGreen}(+4.70\%) \\
\hline
\end{tabular}
}
\label{tab:results}
\end{table*}

To demonstrate the effectiveness of AIM, we conduct experiments on 5 different merging methods under 4 different scenarios. As mentioned before, we use 3 different FT LLM experts in our experiments. As such, we merge models using each merging method for all 4 possible permutations of these expert LLMs. Then we apply AIM to all merged models and measure the performance of each model in all 6 benchmarks. We also report the HV gain for each merged model compared to the population of the base model and the models being merged~(in cases with 2 models, the population will only include the models used for merging). These results are presented in Table \ref{tab:results}. For this experiment, we used $\omega=0.4$, which we found to be the best balance of performance among the various merging methods we use. This choice was informed by our analysis in Section \ref{sec:ablation}. In Table~\ref{tab:results}, we have highlighted the gain/loss of performance for each benchmark due to AIM and we can see that in the vast majority of cases, AIM causes a significant performance boost, with 
an Average Change of ~13\% (ignoring the Inf value) and more than 40\% HV Gain in 20\% cases, further highlighted by the fact that the top performers on each benchmark, as well as the largest hypervolume gain, are all in models merged with AIM. We observe HumanEval (10 out of 20) and MBPP (17 out of 20) often see large boosts with AIM, especially when merging Instruction Tuned models with others. Some merges also reveal small drops in GSM8K or IFEval even when other benchmarks improve, reflecting the inherent trade-offs in merging specialized models.
Overall, a clear majority \textbf{(80\%) of merges exhibit improved HV Gain under AIM}, reinforcing that the method often enhances multi-task performance overall. We can further visualize this increase in hypervolume by looking at how AIM pushes the Pareto frontier.
Figure \ref{fig:frontier} shows how applying AIM to existing merging methods extends the Pareto optimal frontier, which we also quantitatively measured using HV gain.
These results showcase the efficacy of the proposed method across a variety of merging methods and reinforce the hypothesis that the activation space encompasses useful insight for merging.

To validate AIM's generalizability beyond the Llama-2 family, we conducted a new experiment on a different architecture and modality: the Qwen 2.5-VL-7B vision-language model. We merged two distinct experts, Video-R1 model for video reasoning \citep{feng2025videoR1} and CAD-Coder model for image-to-code-based CAD geometry generation \citep{doris2025cadcoder}, with their instruct base model. 

We evaluated this merge on 6 benchmarks: IFEval and MMLU for instruct base model capabilities, Video MMMU \citep{hu2025videommmu} and VSI-Bench \citep{yang2025vsi} for video reasoning, and two CAD-Coder benchmarks for code generation from rendered and real images \citep{doris2025cadcoder}. As shown in Table \ref{table:result-qwen}, AIM consistently improves the underlying merge methods across these diverse tasks and improves the Hypervolume Gain under all four merging methods. This demonstrates that AIM's benefits are not architecture-specific and generalize effectively to multimodal models.

For merging, therefore, we have two experts, one for CAD and one for video reasoning, with a base model that is instruction-tuned. Therefore, to assess performance, we perform the IF-Eval and MMLU benchmarks (the base model knowledge and instruction following) as well as expert benchmarks for video reasoning, Video MMMU \citep{hu2025videommmu} and VSI-Bench \citep{yang2025vsi}, and CAD generation benchmarks of CAD-Coder benchmark (on rendered CAD images) \citep{doris2025cadcoder} and CAD-Coder Real benchmark (on real 3D printed images) \citep{doris2025cadcoder}. As shown in Table \ref{table:result-qwen} below, AIM consistently improves the performance of the underlying merging method across diverse tasks while also showing that the overall hypervolume gain is raised in all merging scenarios, demonstrating AIM's benefits are not limited to the Llama-2 architecture, and AIM does provide a path towards higher quality merging with little computational and data overhead. 

\begin{table*}[ht]
\vspace{-0.5em}
\centering
\caption{\textbf{AIM generalizes across architectures and modalities.} We evaluated merging two Qwen 2.5-VL-7B experts (Video-R1 and CADCoder, with an instruction-tuned base). AIM consistently improves performance across benchmarks, increasing the multi-task Hypervolume Gain in all cases.}
\small
\resizebox{\textwidth}{!}{
\begin{tabular}{l|l|ccccccc|c}
\hline
Method & AIM & IFEval & MMLU & Video VSI & Video MMMU & CAD test100 & CAD R400 & HV Gain \\
\hline
\multicolumn{9}{c}{\textbf{Base Models}} \\
\hline
Instruct & - & \cellcolor{orange!20}{0.68} & \cellcolor{orange!20}{0.67} & 0.21 & 0.47 & 0.04 & 0.05 & - \\
Video-R1 & - & 0.64 & 0.61 & \cellcolor{orange!20}{0.38} & \cellcolor{orange!20}{0.49} & 0.00 & 0.03 & - \\
CADCoder & - & 0.27 & 0.66 & 0.00 & 0.00 & \cellcolor{orange!20}{0.61} & \cellcolor{orange!20}{0.32} & - \\
\hline
\multicolumn{9}{c}{\textbf{Merged Models}} \\
\hline
\multirow{2}{*}{TIES} & No & 0.57 & 0.68 & 0.23 & 0.41 & 0.61 & 0.32 & 0.44 \\
& Yes & 0.58 \color{DarkGreen}(+1.75\%) & 0.68 & 0.23 & 0.45 \color{DarkGreen}(+9.76\%) & \cellcolor{blue!20}{0.65} \color{DarkGreen}(+6.56\%) & 0.32 & \cellcolor{blue!20}{0.45} \color{DarkGreen}(+2.26\%) \\
\hline
\multirow{2}{*}{DARE} & No & 0.54 & 0.67 & 0.25 & 0.42 & 0.54 & 0.32 & 0.43 \\
& Yes & 0.56 \color{DarkGreen}(+3.70\%) & 0.67 & 0.26 \color{DarkGreen}(+4.00\%) & 0.45 \color{DarkGreen}(+7.14\%) & 0.55 \color{DarkGreen}(+1.85\%) & 0.32 & 0.44 \color{DarkGreen}(+3.47\%) \\
\hline
\multirow{2}{*}{DARE Task Arithmetic} & No & 0.25 & 0.64 & 0.25 & 0.43 & 0.54 & 0.33 & 0.38 \\
& Yes & 0.56 \color{DarkGreen}(+124.00\%) & \cellcolor{blue!20}{0.67} \color{DarkGreen}(+4.69\%) & 0.26 \color{DarkGreen}(+4.00\%) & 0.45 \color{DarkGreen}(+4.65\%) & 0.54 & \cellcolor{blue!20}{0.34} \color{DarkGreen}(+3.03\%) & 0.45 \color{DarkGreen}(+17.42\%) \\
\hline
\multirow{2}{*}{WIDEN} & No & 0.35 & 0.66 & 0.08 & 0.26 & 0.61 & 0.33 & 0.32 \\
& Yes & 0.35 & 0.66 & 0.14 \color{DarkGreen}(+75.00\%) & 0.32 \color{DarkGreen}(+23.08\%) & 0.64 \color{DarkGreen}(+4.92\%) & 0.33 & 0.36 \color{DarkGreen}(+13.24\%) \\
\hline
\end{tabular}
}
\label{table:result-qwen}
\vspace{-1.5em}
\end{table*}

\subsection{Ablation study}
 
\label{sec:ablation}
\begin{wrapfigure}{R}{0.45\linewidth}
    \vskip -2em
    \centering
    \includegraphics[width=0.9\linewidth]{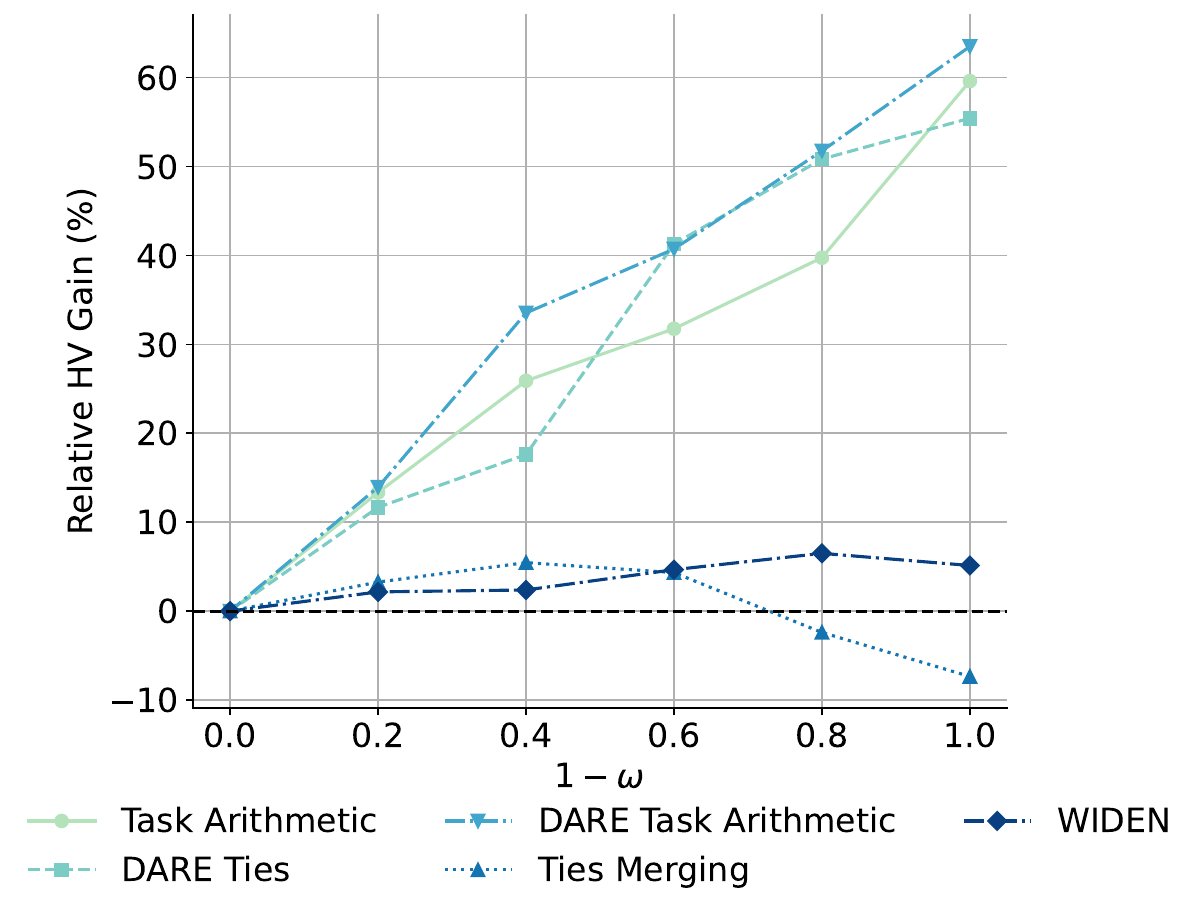}
    \caption{\textbf{The Impact of the Relaxation Factor $\omega$ on Merged Model Performance}. This figure plots the relative change in HV-Gain compared to scenarios without AIM. The x-axis represents $1-\omega$, reflecting that decreasing $\omega$ results in more relaxation. The plot indicates that for some tasks, smaller values of $\omega$ continue to yield benefits. An $\omega$ of 0.4-0.6 appears to strike a balance.}
    \label{fig:ablation}
    \vskip -1em
\end{wrapfigure}

To understand the effects of changing $\omega$ in AIM, we conduct an ablation study on the case of merging all three expert LLMs. For this study, we apply AIM with $\omega \in \{0.0, 0.2, 0.4, 0.6, 0.8\}$ and run the benchmarks on each merged model with each value of $\omega$. For brevity, we do not report all benchmark results for each value here; instead, we track the hypervolume gain~(The full set of results are presented in Appendix \ref{app:ablation}). Specifically, to better visualize the effect of $\omega$, we measure the relative change in HV Gain compared to no AIM~(i.e., $\omega=1.0$). We present these results in Figure \ref{fig:ablation}. In most merging methods, we see that decreasing $\omega$ to even $0$ benefits the model performance. However, in TIES merging particularly, we see that decreasing $\omega$ beyond $0.4$ seems to degrade performance, and setting $\omega$ to the most extreme case of $0.0$ does see some degradation in WIDEN as well. Given this, it seems that in these experiments, a value of $0.4$ balances the performance gains in methods responding well to AIM and the potential degradation of methods that benefit less from AIM. However, given this observation that in some cases pushing $\omega$ to $0$ still yields benefits, there may be some value in exploring non-linear scaling of activation magnitudes and non-linear relaxation schemes that could further boost performance in some cases.

\begin{wrapfigure}{R}{0.5\linewidth}
    \vskip -25em
    \centering
    \includegraphics[width=\linewidth]
    {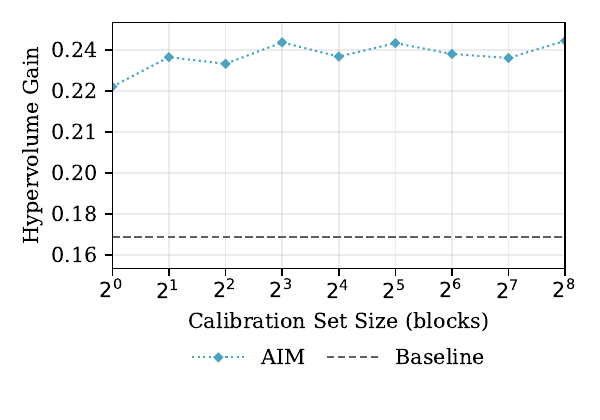}
    \vskip -2em
    \caption{\textbf{AIM's Robustness to Calibration Set Size}. HV-Gain is plotted against the number of calibration blocks (log scale) for the DARE TIES merge. The dashed line is the baseline performance without AIM. Significant improvement is achieved with small data, and performance stabilizes at only 8 blocks.}
    \label{fig:calibration_sensitivity}
\end{wrapfigure}

To empirically validate the sensitivity to the calibration set, we conducted an ablation study on the calibration set size. We applied AIM to the DARE TIES merge of all three Llama-2 experts from Table \ref{tab:results}, varying the number of calibration blocks from 1 to 256. As shown in Figure \ref{fig:calibration_sensitivity}, the results demonstrate that AIM is highly robust. The Hypervolume Gain (HV Gain) increases substantially with just one block and stabilizes by 8 blocks (sourced from the Pile corpus \citet{gao2020pile800gbdatasetdiverse}). This confirms that the data overhead for AIM is minimal and the method is highly practical.

\section{Conclusion and Outlook}
 \vspace{-1em}
\label{sec:conclusion}
In this work, we introduced Activation-Informed Merging (AIM) as a complementary algorithm to existing model merging techniques for large language models (LLMs). We hypothesized that the activation space of LLMs harbors useful information that is often overlooked in model merging, as most existing methods operate purely on the weight space. To explore this potential information in the activation space, we viewed the problem from a continual learning perspective and proposed leveraging the activation space information from a task-agnostic calibration set. This approach selectively preserves critical weights from the pre-trained model, mitigating catastrophic forgetting while incorporating knowledge from fine-tuned models, yielding overall higher-performing models.

Through extensive empirical evaluations across multiple merging methods and benchmark tasks, and model architectures (including the Llama-2 and Qwen-VL families), we demonstrated that AIM consistently improves performance, often yielding superior results in comparison to the original merging methods it was applied to. These results empirically confirm our hypothesis on the importance of the activation space. Notably, AIM boosted merged model performance by up to \textbf{40\%} in some cases, underscoring the crucial role and the potential of activation information in merging methods. Furthermore, our ablation study confirmed AIM's robustness, showing significant gains even with minimal calibration data (as few as 8 blocks), highlighting the method's practical applicability. Our findings strongly highlight the necessity and benefit of incorporating activation-informed strategies when merging multiple fine-tuned models.

Moving forward, our findings open up several promising directions for future research. First, our results indicate that even aggressively preserving salient weights of the pre-trained model is effective across many merging scenarios. This highlights the promise for more advanced activation-informed strategies and non-linear relaxation methods to potentially further enhance performance. Beyond the pre-trained activations explored in this work,  there is room to improve existing merging methods by leveraging the broader activation space of the models being merged. So far, AIM has only considered the activations of the pre-trained model, while the activations of the expert LLMs remain unexplored. Future research should focus on developing methods that also encompass information from the expert model activations. Additionally, in future works, more theoretically grounded approaches for incorporating the activation space of LLMs in merging should be developed and tested. These integrations will hold great value in improving the quality and performance of merging methods in an increasingly competitive and ever more efficient landscape of LLMs, which could benefit from smaller and more efficient yet more powerful models.

Overall, AIM serves as a robust and adaptable augmentation to existing LLM merging techniques, offering a principled way to incorporate activation information for more effective model fusion. By prioritizing the activation-aware perspective, we take a step towards more stable, efficient, and generalizable merged models, demonstrated across different architectures and modalities, that better leverage the strengths of multiple fine-tuned experts.
\vspace{-1em}

\begin{ack}
\vspace{-1em}
 This work was supported in part by the MIT-IBM Watson AI Lab.
\end{ack}

\bibliography{ref}
\bibliographystyle{cleancite}

\clearpage
\appendix
\section{Reproducibility Details}
\label{app:reporducibility}
Here we provide the specific details for reproducing the results presented in the paper. 

\paragraph{Checkpoints:} Firstly, we specify the publicly available checkpoints we use in our experiments. Below is a list of checkpoints used and the links to the publicly available weights for these models:

\begin{itemize}
    \item \textbf{Base Model: } \href{https://huggingface.co/unsloth/llama-2-13b}{https://huggingface.co/unsloth/llama-2-13b}
    \item \textbf{Code Model: } \href{https://huggingface.co/layoric/llama-2-13b-code-alpaca}{https://huggingface.co/layoric/llama-2-13b-code-alpaca}
    \item \textbf{Math Model: }\href{https://huggingface.co/vanillaOVO/WizardMath-13B-V1.0}{https://huggingface.co/vanillaOVO/WizardMath-13B-V1.0}
    \item \textbf{Instruction Tuned Model: }\href{https://huggingface.co/WizardLMTeam/WizardLM-13B-V1.2}{https://huggingface.co/WizardLMTeam/WizardLM-13B-V1.2}
\end{itemize}

\textbf{A Note On Weights: } The weights we use in our experiments may not be exactly identical to the weights used in the experiments by \citet{DARE} and \citet{widen}, since the referenced weights for WizardMath-13B are no longer available publicly, instead we use a publicly available copy of the model.

\paragraph{Code and Data:} Aside from the checkpoint, we provide our code and the link to the publicly available calibration data we use in our work. Our code is publicly available at \href{https://github.com/ahnobari/ActivationInformedMerging}{https://github.com/ahnobari/ActivationInformedMerging} and the calibration data can be found at \href{https://huggingface.co/datasets/mit-han-lab/pile-val-backup}{https://huggingface.co/datasets/mit-han-lab/pile-val-backup}.

\clearpage
\section{Ablation Detailed Results}
\label{app:ablation}
Here we present the full results of the ablation study we conducted. Table \ref{tab:fullablation} includes the granular values for all benchmarks we ran for different values of $\omega$.

\begin{table}[h]
    \centering
    \caption{Performance metrics for different methods with varying $\omega$ values.}
    \resizebox{!}{0.4\linewidth}{
    \begin{tabular}{l c c c c c c c c}
        \toprule
        Method & $\omega$ & HumanEval & MBPP & MMLU & MATH & GSM8K & IFEval & HV Gain \\
        \midrule
        \multirow{6}{*}{Ties Merging} 
        & 0.0 & 19.51 & 29.0 & 54.32 & 4.6 & 17.6 & 26.24 & 0.1015 \\
        & 0.2 & 19.51 & 28.6 & 54.35 & 5.3 & 21.2 & 25.37 & 0.1069 \\
        & 0.4 & 20.73 & 29.2 & 54.46 & 5.7 & 23.7 & 25.98 & 0.1143 \\
        & 0.6 & 20.12 & 27.8 & 54.25 & 6.6 & 32.1 & 23.9 & 0.1155 \\
        & 0.8 & 20.73 & 27.8 & 54.09 & 6.6 & 33.7 & 24.07 & 0.1131 \\
        & 1.0 & 21.34 & 29.2 & 53.97 & 6.3 & 29.2 & 26.95 & 0.1096 \\
        \midrule
        \multirow{6}{*}{DARE Task Arithmetic} 
        & 0.0 & 18.90 & 29.4 & 53.42 & 13.8 & 60.5 & 35.49 & 0.2623 \\
        & 0.2 & 16.46 & 29.0 & 52.98 & 12.9 & 61.5 & 34.97 & 0.2434 \\
        & 0.4 & 15.85 & 27.0 & 52.59 & 12.2 & 60.7 & 33.67 & 0.2257 \\
        & 0.6 & 15.24 & 27.0 & 52.18 & 11.8 & 58.1 & 33.95 & 0.2142 \\
        & 0.8 & 14.02 & 22.2 & 51.53 & 9.9 & 54.0 & 32.69 & 0.1828 \\
        & 1.0 & 11.59 & 19.6 & 50.89 & 9.1 & 49.7 & 33.2 & 0.1604 \\
        \midrule
        \multirow{6}{*}{DARE Ties} 
        & 0.0 & 25.61 & 29.6 & 53.31 & 12.0 & 59.4 & 34.64 & 0.2669 \\
        & 0.2 & 21.34 & 29.4 & 53.11 & 12.1 & 58.7 & 36.76 & 0.2590 \\
        & 0.4 & 19.51 & 28.6 & 52.63 & 11.6 & 57.0 & 36.2 & 0.2426 \\
        & 0.6 & 15.85 & 26.0 & 52.22 & 10.1 & 54.2 & 34.82 & 0.2019 \\
        & 0.8 & 14.02 & 24.0 & 51.6 & 10.0 & 53.1 & 35.51 & 0.1917 \\
        & 1.0 & 13.41 & 21.2 & 51.15 & 8.7 & 51.5 & 35.75 & 0.1717 \\
        \midrule
        \multirow{6}{*}{Task Arithmetic} 
        & 0.0 & 18.90 & 29.4 & 53.42 & 13.8 & 60.5 & 35.49 & 0.2623 \\
        & 0.2 & 15.24 & 27.6 & 52.97 & 13.0 & 59.8 & 35.27 & 0.2296 \\
        & 0.4 & 15.24 & 27.4 & 52.63 & 12.0 & 58.1 & 33.88 & 0.2165 \\
        & 0.6 & 15.24 & 25.4 & 52.13 & 11.4 & 57.4 & 33.29 & 0.2069 \\
        & 0.8 & 13.41 & 21.8 & 51.61 & 10.2 & 56.2 & 32.74 & 0.1862 \\
        & 1.0 & 11.59 & 19.6 & 51.20 & 9.0 & 52.7 & 32.95 & 0.1643 \\
        \midrule
        \multirow{6}{*}{WIDEN} 
        & 0.0 & 27.44 & 33.2 & 55.26 & 14.0 & 64.9 & 32.39 & 0.3027 \\
        & 0.2 & 28.05 & 33.0 & 55.16 & 14.2 & 65.6 & 32.39 & 0.3066 \\
        & 0.4 & 26.83 & 32.8 & 54.98 & 14.4 & 64.0 & 32.76 & 0.3013 \\
        & 0.6 & 25.61 & 33.4 & 54.77 & 14.2 & 63.0 & 32.06 & 0.2947 \\
        & 0.8 & 26.22 & 32.6 & 54.64 & 14.0 & 64.1 & 31.74 & 0.2941 \\
        & 1.0 & 25.00 & 33.2 & 54.58 & 13.5 & 64.2 & 31.44 & 0.2879 \\
        \bottomrule
    \end{tabular}
    }
    \label{tab:fullablation}
\end{table}

\clearpage
\section{Sensitivity-Based Formulation Comparison}
\label{app:continualexp}
In this section, we present the results of running post-merging relaxation using the gradient-based sensitivity-based formulation discussed in the main body. We run the relaxation scheme based on gradients with a $\omega=0.4$, which we use in AIM. Table \ref{tab:continual} shows how performance changes across benchmarks with gradient-based relaxation. We observe that the resulting performance boost remains very close to activation-based relaxation, with HV gain largely unchanged, with both gradient-based and pure activation-based boost trading blows evenly (see Table \ref{tab:results}). Noting this and given that activations alone do not come with significant memory and computational cost, activation-based relaxation, which does not require computing gradients, is a more memory-efficient and computationally inexpensive process.

\begin{table}[htbp]
\centering
\caption{Model performance comparison across different benchmarks, after relaxation applied using the gradients of the models with respect to the calibration data.}
\small
\resizebox{\textwidth}{!}{
\begin{tabular}{l|l|l|ccccccc}
\hline
Method & Model(s) & Relaxation & HumanEval & MBPP & MMLU & MATH & GSM8K & IFEval & HV Gain \\
\hline
\multicolumn{9}{c}{\textbf{Base Models}} \\
\hline
- & Base & - & 17.07 & 27.80 & 52.18 & 0.70 & 4.20 & 25.10 & - \\
- & Code & - & 17.07 & 31.60 & 52.91 & 6.00 & 24.10 & 26.25 & - \\
- & Instruction Tuned & - & \cellcolor{orange!20}{26.83} & \cellcolor{orange!20}{34.80} & \cellcolor{orange!20}{53.41} & 7.50 & 43.40 & \cellcolor{orange!20}{35.67} & - \\
- & Math & - & 15.24 & 27.60 & 51.89 & \cellcolor{orange!20}{13.10} & \cellcolor{orange!20}{59.10} & 21.58 & - \\
\hline
\multicolumn{9}{c}{\textbf{Merged Models}} \\
\hline
\multirow{8}{*}{DARE Average} & \multirow{2}{*}{Code + Instruction Tuned} & No & 26.83 & 34.40 & 53.53 & 8.40 & 45.80 & 33.42 & 0.27 \\
& & Yes & 29.27 \color{DarkGreen}(+9.09\%) & 36.00 \color{DarkGreen}(+4.65\%) & 54.18 \color{DarkGreen}(+1.21\%) & 8.30 \color{DarkRed}(-1.19\%) & 46.20 \color{DarkGreen}(+0.87\%) & 32.00 \color{DarkRed}(-4.25\%) & 0.28 \color{DarkGreen}(+2.49\%) \\
 & \multirow{2}{*}{Code + Math} & No & 16.46 & 28.60 & 51.96 & 15.10 & 64.70 & 22.02 & 0.23 \\
& & Yes & 15.85 \color{DarkRed}(-3.71\%) & 29.60 \color{DarkGreen}(+3.50\%) & 52.50 \color{DarkGreen}(+1.04\%) & 14.80 \color{DarkRed}(-1.99\%) & 64.10 \color{DarkRed}(-0.93\%) & 21.91 \color{DarkRed}(-0.50\%) & 0.23 \color{DarkRed}(-1.65\%) \\
 & \multirow{2}{*}{Instruction Tuned + Math} & No & 5.49 & 19.00 & 51.08 & 9.80 & 54.30 & 32.35 & 0.18 \\
& & Yes & 12.20 \color{DarkGreen}(+122.22\%) & 28.20 \color{DarkGreen}(+48.42\%) & 52.72 \color{DarkGreen}(+3.21\%) & 12.90 \color{DarkGreen}(+31.63\%) & 62.20 \color{DarkGreen}(+14.55\%) & 31.96 \color{DarkRed}(-1.21\%) & 0.26 \color{DarkGreen}(+40.71\%) \\
 & \multirow{2}{*}{Code + Instruction Tuned + Math} & No & 11.59 & 19.60 & 50.89 & 9.10 & 49.70 & 33.20 & 0.16 \\
& & Yes & 15.85 \color{DarkGreen}(+36.76\%) & 27.00 \color{DarkGreen}(+37.76\%) & 52.59 \color{DarkGreen}(+3.34\%) & 12.20 \color{DarkGreen}(+34.07\%) & 60.70 \color{DarkGreen}(+22.13\%) & 33.59 \color{DarkGreen}(+1.17\%) & 0.23 \color{DarkGreen}(+40.59\%) \\
\hline
\multirow{8}{*}{DARE Ties} & \multirow{2}{*}{Instruction Tuned + Math} & No & 8.54 & 23.80 & 51.39 & 9.20 & 54.10 & 33.89 & 0.20 \\
& & Yes & 15.85 \color{DarkGreen}(+85.60\%) & 30.20 \color{DarkGreen}(+26.89\%) & 52.89 \color{DarkGreen}(+2.92\%) & 11.60 \color{DarkGreen}(+26.09\%) & 57.80 \color{DarkGreen}(+6.84\%) & 35.63 \color{DarkGreen}(+5.13\%) & 0.26 \color{DarkGreen}(+31.22\%) \\
 & \multirow{2}{*}{Code + Instruction Tuned} & No & \cellcolor{blue!20}{30.49} & 35.20 & 53.40 & 8.60 & 46.20 & 33.28 & 0.28 \\
& & Yes & \cellcolor{blue!20}{30.49} & \cellcolor{blue!20}{36.80} \color{DarkGreen}(+4.55\%) & 54.02 \color{DarkGreen}(+1.16\%) & 8.60 & 47.20 \color{DarkGreen}(+2.16\%) & 33.16 \color{DarkRed}(-0.36\%) & 0.29 \color{DarkGreen}(+1.63\%) \\
 & \multirow{2}{*}{Code + Math} & No & 17.07 & 27.40 & 51.92 & 14.90 & 63.60 & 22.53 & 0.23 \\
& & Yes & 17.68 \color{DarkGreen}(+3.57\%) & 29.00 \color{DarkGreen}(+5.84\%) & 52.61 \color{DarkGreen}(+1.33\%) & 15.20 \color{DarkGreen}(+2.01\%) & 63.90 \color{DarkGreen}(+0.47\%) & 21.10 \color{DarkRed}(-6.35\%) & 0.24 \color{DarkGreen}(+4.00\%) \\
 & \multirow{2}{*}{Code + Instruction Tuned + Math} & No & 13.41 & 21.20 & 51.15 & 8.70 & 51.50 & 35.75 & 0.17 \\
& & Yes & 19.51 \color{DarkGreen}(+45.49\%) & 28.60 \color{DarkGreen}(+34.91\%) & 52.63 \color{DarkGreen}(+2.89\%) & 11.60 \color{DarkGreen}(+33.33\%) & 57.00 \color{DarkGreen}(+10.68\%) & \cellcolor{blue!20}{36.20} \color{DarkGreen}(+1.26\%) & 0.24 \color{DarkGreen}(+41.28\%) \\
\hline
\multirow{8}{*}{Task Arithmetic} & \multirow{2}{*}{Code + Instruction Tuned} & No & 29.27 & 33.80 & 53.44 & 8.60 & 47.10 & 31.60 & 0.28 \\
& & Yes & 29.88 \color{DarkGreen}(+2.08\%) & 35.80 \color{DarkGreen}(+5.92\%) & 54.12 \color{DarkGreen}(+1.27\%) & 7.80 \color{DarkRed}(-9.30\%) & 46.60 \color{DarkRed}(-1.06\%) & 32.01 \color{DarkGreen}(+1.30\%) & 0.28 \color{DarkGreen}(+0.61\%) \\
 & \multirow{2}{*}{Instruction Tuned + Math} & No & 4.27 & 20.20 & 51.50 & 10.00 & 54.20 & 31.31 & 0.18 \\
& & Yes & 8.54 \color{DarkGreen}(+100.00\%) & 26.40 \color{DarkGreen}(+30.69\%) & 52.83 \color{DarkGreen}(+2.58\%) & 12.80 \color{DarkGreen}(+28.00\%) & 61.30 \color{DarkGreen}(+13.10\%) & 32.62 \color{DarkGreen}(+4.18\%) & 0.24 \color{DarkGreen}(+34.52\%) \\
 & \multirow{2}{*}{Code + Math} & No & 18.29 & 28.60 & 52.10 & 15.00 & 64.70 & 21.92 & 0.24 \\
& & Yes & 17.68 \color{DarkRed}(-3.34\%) & 29.20 \color{DarkGreen}(+2.10\%) & 52.52 \color{DarkGreen}(+0.81\%) & 14.60 \color{DarkRed}(-2.67\%) & 64.50 \color{DarkRed}(-0.31\%) & 21.54 \color{DarkRed}(-1.73\%) & 0.24 \color{DarkRed}(-2.65\%) \\
 & \multirow{2}{*}{Code + Instruction Tuned + Math} & No & 11.59 & 19.60 & 51.20 & 9.00 & 52.70 & 32.87 & 0.16 \\
& & Yes & 15.24 \color{DarkGreen}(+31.49\%) & 27.40 \color{DarkGreen}(+39.80\%) & 52.63 \color{DarkGreen}(+2.79\%) & 12.00 \color{DarkGreen}(+33.33\%) & 58.10 \color{DarkGreen}(+10.25\%) & 33.91 \color{DarkGreen}(+3.16\%) & 0.22 \color{DarkGreen}(+31.97\%) \\
\hline
\multirow{8}{*}{Ties Merging} & \multirow{2}{*}{Code + Math} & No & 15.85 & 26.80 & 51.86 & 14.30 & 62.60 & 21.63 & 0.20 \\
& & Yes & 15.85 & 28.60 \color{DarkGreen}(+6.72\%) & 52.29 \color{DarkGreen}(+0.83\%) & \cellcolor{blue!20}{15.30} \color{DarkGreen}(+6.99\%) & 63.80 \color{DarkGreen}(+1.92\%) & 22.64 \color{DarkGreen}(+4.67\%) & 0.23 \color{DarkGreen}(+13.55\%) \\
 & \multirow{2}{*}{Instruction Tuned + Math} & No & 28.05 & 34.60 & 54.45 & 8.70 & 44.70 & 34.04 & 0.23 \\
& & Yes & 27.44 \color{DarkRed}(-2.17\%) & 35.00 \color{DarkGreen}(+1.16\%) & 54.74 \color{DarkGreen}(+0.53\%) & 9.30 \color{DarkGreen}(+6.90\%) & 46.10 \color{DarkGreen}(+3.13\%) & 34.51 \color{DarkGreen}(+1.38\%) & 0.25 \color{DarkGreen}(+6.38\%) \\
 & \multirow{2}{*}{Code + Instruction Tuned} & No & 16.46 & 23.60 & 52.70 & 2.70 & 5.40 & 24.48 & 0.00 \\
& & Yes & 15.24 \color{DarkRed}(-7.41\%) & 24.20 \color{DarkGreen}(+2.54\%) & 53.15 \color{DarkGreen}(+0.85\%) & 2.60 \color{DarkRed}(-3.70\%) & 5.20 \color{DarkRed}(-3.70\%) & 22.87 \color{DarkRed}(-6.58\%) & 0.05 \color{DarkGreen}(+inf\%) \\
 & \multirow{2}{*}{Code + Instruction Tuned + Math} & No & 21.34 & 29.20 & 53.97 & 6.30 & 29.20 & 26.95 & 0.11 \\
& & Yes & 20.73 \color{DarkRed}(-2.86\%) & 29.20 & 54.46 \color{DarkGreen}(+0.91\%) & 5.70 \color{DarkRed}(-9.52\%) & 23.70 \color{DarkRed}(-18.84\%) & 25.98 \color{DarkRed}(-3.60\%) & 0.11 \color{DarkGreen}(+4.33\%) \\
\hline
\multirow{8}{*}{WIDEN} & \multirow{2}{*}{Instruction Tuned + Math} & No & 24.39 & 30.40 & 54.20 & 14.60 & 66.00 & 30.82 & 0.30 \\
& & Yes & 23.78 \color{DarkRed}(-2.50\%) & 32.00 \color{DarkGreen}(+5.26\%) & 54.69 \color{DarkGreen}(+0.90\%) & 15.10 \color{DarkGreen}(+3.42\%) & \cellcolor{blue!20}{68.20} \color{DarkGreen}(+3.33\%) & 31.23 \color{DarkGreen}(+1.33\%) & \cellcolor{blue!20}{0.31} \color{DarkGreen}(+2.54\%) \\
 & \multirow{2}{*}{Code + Math} & No & 17.07 & 29.40 & 53.35 & 14.20 & 64.40 & 24.02 & 0.24 \\
& & Yes & 17.07 & 29.60 \color{DarkGreen}(+0.68\%) & 53.36 \color{DarkGreen}(+0.02\%) & 14.30 \color{DarkGreen}(+0.70\%) & 62.20 \color{DarkRed}(-3.42\%) & 23.95 \color{DarkRed}(-0.29\%) & 0.24 \color{DarkRed}(-1.22\%) \\
 & \multirow{2}{*}{Code + Instruction Tuned} & No & 26.22 & 35.60 & 54.90 & 8.30 & 45.00 & 30.42 & 0.27 \\
& & Yes & 25.61 \color{DarkRed}(-2.33\%) & 34.60 \color{DarkRed}(-2.81\%) & 54.97 \color{DarkGreen}(+0.13\%) & 8.20 \color{DarkRed}(-1.20\%) & 44.10 \color{DarkRed}(-2.00\%) & 31.60 \color{DarkGreen}(+3.88\%) & 0.26 \color{DarkRed}(-0.93\%) \\
 & \multirow{2}{*}{Code + Instruction Tuned + Math} & No & 25.00 & 33.20 & 54.58 & 13.50 & 64.20 & 31.44 & 0.29 \\
& & Yes & 26.83 \color{DarkGreen}(+7.32\%) & 32.80 \color{DarkRed}(-1.20\%) & \cellcolor{blue!20}{54.98} \color{DarkGreen}(+0.73\%) & 14.40 \color{DarkGreen}(+6.67\%) & 64.00 \color{DarkRed}(-0.31\%) & 32.82 \color{DarkGreen}(+4.39\%) & 0.30 \color{DarkGreen}(+4.70\%) \\
\hline
\end{tabular}
}
\label{tab:continual}
\end{table}

\end{document}